\documentclass{article}

\PassOptionsToPackage{numbers, compress}{natbib}
\usepackage[final]{nips_2017}

\usepackage{graphicx}
\usepackage[usenames,dvipsnames]{color}

\usepackage[utf8]{inputenc} 
\usepackage[T1]{fontenc}    
\usepackage{hyperref}       
\usepackage{url}            
\usepackage{booktabs}       
\usepackage{amsfonts}       
\usepackage{nicefrac}       
\usepackage{microtype}      
\usepackage[ruled,vlined,linesnumbered]{algorithm2e}
\usepackage{amsmath}
\usepackage{bbm}
\usepackage{hyperref,caption}
\usepackage[]{todonotes}
\usepackage{caption}
\usepackage{subcaption}
\usepackage{graphicx}
\usepackage{wrapfig}

\usepackage{enumitem}

\title{CatBoost: gradient boosting with categorical features support}

\newcommand{\X}{\mathbf{X}}
\newcommand{\D}{\mathcal{D}}
\newcommand{\R}{\mathbb{R}}

\author{
  Anna Veronika  Dorogush, Vasily Ershov, Andrey  Gulin \\
  Yandex \\
}

\begin{document}

\maketitle

\begin{abstract}
In this paper we present CatBoost, a new open-sourced gradient boosting library that successfully handles categorical features and outperforms existing publicly available implementations of gradient boosting in terms of quality on a set of popular publicly available datasets. The library has a GPU implementation of learning algorithm and a CPU implementation of scoring algorithm, which are significantly faster than other gradient boosting libraries on ensembles of similar sizes.
\end{abstract}

\section{Introduction}\label{sec:introduction}

Gradient boosting is a powerful machine-learning technique that achieves state-of-the-art results in a variety of practical tasks. For a number of years, it has remained the primary method for learning problems with heterogeneous features, noisy data, and complex dependencies: web search, recommendation systems, weather forecasting, and many others~\cite{caruana2006empirical,roe2005boosted,wu2010adapting,zhang2015gradient}.  It is backed by strong 
theoretical results that explain how strong predictors can be built by iterative combining weaker models (\textit{base predictors}) via a greedy procedure that corresponds to gradient descent in a function space.

Most popular implementations of gradient boosting use decision trees as base predictors.
It is convenient to use decision trees for numerical features, but, in practice, many datasets include categorical features, which are also important for prediction. Categorical feature is a feature having a discrete set of values that are not necessary comparable with each other (e.g., user ID or name of a city). The most commonly used practice for dealing with categorical features in gradient boosting is converting them to numbers before training.

In this paper we present a new gradient boosting algorithm that successfully handles categorical features and takes advantage of dealing with them during training as opposed to preprocessing time.
Another advantage of the algorithm is that it uses a new schema for calculating leaf values when selecting the tree structure, which helps to reduce overfitting.

As a result, the new algorithm outperforms the existing state-of-the-art implementations of gradient boosted decision trees (GBDTs) XGBoost~\cite{chen2016xgboost},  LightGBM\footnote{\url{https://github.com/Microsoft/LightGBM}} and H2O\footnote{\url{http://docs.h2o.ai/h2o/latest-stable/h2o-docs/data-science/gbm.html}}, on a diverse set of popular tasks (Sec.~\ref{Sec:Experiments}).
The algorithm is called CatBoost (for ``categorical boosting'') and is released in open source.\footnote{\url{https://github.com/catboost/catboost}}

CatBoost has both CPU and GPU implementations. The GPU implementation allows for much faster training and is faster than both state-of-the-art open-source GBDT GPU implementations, XGBoost and LightGBM, on ensembles of similar sizes.
The library also has a fast CPU scoring implementation, which outperforms XGBoost and LightGBM implementations on ensembles of similar sizes.

\section{Categorical features}
\label{sec:cat_features}
Categorical features have a discrete set of values called \textit{categories} which are not necessary comparable with each other; thus, such features cannot be used in binary decision trees directly.
A common practice for dealing with categorical features is converting them to numbers at the preprocessing time, i.e., each category for each example is substituted with one or several numerical values.

The most widely used technique which is usually applied to low-cardinality categorical features is \textit{one-hot encoding}: the original feature is removed and a new binary variable is added for each category~\cite{micci2001preprocessing}.
One-hot encoding can be done during the preprocessing phase or during training, the latter can be implemented more efficiently in terms of training time and is implemented in CatBoost. 

Another way to deal with categorical features is to compute some statistics using the label values of the examples.
Namely, assume that we are given a dataset of observations $\D=\{(\X_i,Y_i)\}_{i=1..n}$, where $\X_i=(x_{i,1},\ldots,x_{i,m})$ is a vector of $m$ features, some numerical, some categorical, and $Y_i\in \R$ is a {\it label value}.
The simplest way is to substitute the category with the average label value on the whole train dataset. So, $x_{i,k}$ is substituted with 
$
\frac{\sum_{j=1}^n{[x_{j,k}=x_{i,k}]\cdot Y_j}}{\sum_{j=1}^n[x_{j,k}=x_{i,k}]}\,,
$
where $[\cdot]$ denotes Iverson brackets, i.e., $[x_{j,k}=x_{i,k}]$ equals 1 if $x_{j,k}=x_{i,k}$ and 0 otherwise.
This procedure, obviously, leads to overfitting. For example, if there is a single example from the category $x_{i,k}$ in the whole dataset then the new numeric feature value will be equal to the label value on this example.
A straightforward way to overcome the problem is to partition the dataset into two parts and use one part only to calculate the statistics and the second part to perform training. This reduces overfitting but it also reduces the amount of data used to train the model and to calculate the statistics.

CatBoost uses a more efficient strategy~\cite{dorogush2017fighting} which reduces overfitting and allows to use the whole dataset for training. Namely, we perform a random permutation of the dataset and for each example we compute average label value for the example with the same category value placed before the given one in the permutation.
Let $\sigma=(\sigma_1, \ldots, \sigma_n)$ be the permutation, then $x_{\sigma_p,k}$ is substituted with
\vspace{-6.3pt}
\begin{equation}\label{eq:ordered_ctr}
\frac{\sum_{j=1}^{p-1}{[x_{\sigma_j,k}=x_{\sigma_p,k}]Y_{\sigma_j}} + a\cdot P} {\sum_{j=1}^{p-1}[x_{\sigma_j,k}=x_{\sigma_p,k}] + a},
\vspace{-5.5pt}
\end{equation}
where we also add a prior value $P$ and a parameter $a>0$, which is the weight of the prior.
Adding prior is a common practice and it helps to reduce the noise obtained from low-frequency categories~\cite{cestnik1990estimating}. 
For regression tasks standard technique for calculating prior is to take the average label value in the dataset. For binary classification task a prior is usually an a priori probability of encountering a positive class~\cite{micci2001preprocessing}.
It is also efficient to use several permutations. However, one can see that a straightforward usage of statistics computed for several permutations would lead to overfitting. As we discuss in the next section, CatBoost uses a novel schema for calculating leaf values which allows to use several permutations without this problem.

\textbf{Feature combinations}
Note that any combination of several categorical features could be considered as a new one.
For example, assume that the task is music recommendation and we have two categorical features: user ID and musical genre.
Some user prefers, say, rock music. When we convert user ID and musical genre to numerical features according to~\eqref{eq:ordered_ctr}, we loose this information. A combination of two features solves this problem and gives a new powerful feature.  
However, the number of combinations grows exponentially with the number of categorical features in dataset and it is not possible to consider all of them in the algorithm. When constructing a new split for the current tree, CatBoost considers combinations in a greedy way.
No combinations are considered for the first split in the tree. For the next splits CatBoost combines all combinations and categorical features present in current tree with all categorical features in dataset. Combination values are converted to numbers on the fly. CatBoost also generates combinations of numerical and categorical features in the following way: all the splits selected in the tree are considered as categorical with two values and used in combinations in the same way as categorical ones.

\textbf{Important implementation details}
Another way of substituting category with a number is calculating number of appearances of this category in the dataset. This is a simple but powerful technique and it is implemented in CatBoost. This type of statistic is also calculated for feature combinations. 

In order to fit the optimal prior at each step of CatBoost algorithm, we consider several priors and construct a feature for each of them, which is more efficient in terms of quality than standard techniques mentioned above.

\section{Fighting Gradient Bias}
\label{sec:fighting}

CatBoost, as well as all standard gradient boosting implementations, builds each new tree to approximate the gradients of the current model. However, all classical boosting algorithms suffer from overfitting caused by the problem of biased pointwise gradient estimates.
Gradients used at each step are estimated using the same data points the current model was built on. This leads to a shift of the distribution of estimated gradients in any domain of feature space in comparison with the true distribution of gradients in this domain, which leads to overfitting.
The idea of biased gradients was discussed in previous literature~\cite{breiman1996out}~\cite{friedman2002stochastic}.
We have provided a formal analysis of this problem in the paper~\cite{dorogush2017fighting}. The paper also contains modifications of classical gradient boosting algorithm that try to solve this problem. CatBoost implements one of those modifications, briefly described below.

In many GBDTs (e.g., XGBoost, LightGBM) building next tree comprises two steps: choosing the tree structure and setting values in leafs after the tree structure is fixed.
To choose the best tree structure, the algorithm enumerates through different splits, builds trees with these splits, sets values in the obtained leafs, scores the trees and selects the best split. Leaf values in both phases are calculated as
approximations for gradients~\cite
{friedman2001greedy} or for Newton steps.
In CatBoost the second phase is performed using traditional
GBDT scheme and for the first phase we use the modified version.

According to intuition obtained from our empirical results and our theoretical analysis in~\cite{dorogush2017fighting}, it is highly desirable to use unbiased estimates of the gradient step.
Let $F^{i}$ be the model constructed after building first $i$ trees, $g^{i}(\X_k,Y_k)$ be the gradient value on $k$-th training sample after building $i$ trees.
To make the gradient $g^{i}(\X_k,Y_k)$ unbiased w.r.t. the model $F^{i}$, we need to have $F^{i}$ trained without the observation $\X_k$. Since we need unbiased gradients for all training examples, no observations may be used for training $F^{i}$, which at first glance makes the training process impossible. We consider the following trick to deal with this problem: for each example $\X_k$, we train a separate model $M_k$ that is never updated using a gradient estimate for this example. With $M_k$, we estimate the gradient on $\X_k$ and use this estimate to score the resulting tree.
Let us present the pseudo-code that explains how this trick can be performed.
Let $\text{Loss}(y, a)$ be the optimizing loss function, where $y$ is the label value and $a$ is the formula value. 
\begin{algorithm}[H]
\small
\SetKwInOut{Input}{input}\SetKwInOut{Output}{output}
	\Input{\,\,$\{(\X_k,Y_k)\}_{k=1}^n$ ordered according to $\sigma$, the number of trees $I$\;}
    \BlankLine
    $M_i \leftarrow 0$ for $i = 1..n$\;
            \vspace{0.4mm}
	\For{$iter \leftarrow 1$ \KwTo $I$}{ 
    \vspace{0.3mm}
        \For{$i \leftarrow 1$ \KwTo $n$}{
        \vspace{0.3mm}
        	\For{$j \leftarrow 1$ \KwTo $i-1$}{
            \vspace{0.3mm}
 	            $g_j \leftarrow \frac{d}{da}Loss(y_j, a) |_{a=M_{i}(\X_j)}$\;
          }
          $M \leftarrow LearnOneTree((\X_j,g_j)$ for $j = 1..i-1)$\;
          $M_i \leftarrow M_i + M$\;
          } 
        }
        \Return{$M_1\, \ldots\,M_n; M_1(\X_1), M_2(\X_2)\, M_n(\X_n)$}
	{\caption{Updating the models and calculating model values for gradient estimation}
    \label{alg:ordered}}
\end{algorithm}

Note that $M_i$ is trained without using the example $\X_i$. CatBoost implementation uses the following relaxation of this idea: all $M_i$ share the same tree structures.

In CatBoost we generate $s$ random permutations of our training dataset. We use several permutations to enhance the robustness of the algorithm: we sample a random permutation and obtain gradients on its basis. These are the same permutations as ones used for calculating statistics for categorical features. We use different permutations for training distinct models, thus using several permutations does not lead to overfitting. For each permutation $\sigma$, we train $n$ different models $M_i$, as shown above. That means that for building one tree we need to store and recalculate $O(n^2)$ approximations for each permutation $\sigma$: for each model $M_i$, we have to update $M_i(\X_1), \ldots, M_i(\X_{i}) $. 
Thus, the resulting complexity of this operation is $O(s\,n^2)$.
In our practical implementation, we use one important trick which reduces the complexity of one tree construction to $O(s \, n)$: for each permutation, 
instead of storing and updating $O(n^2)$ values $M_{i}(X_j)$, we maintain values $M'_{i}(\X_j), i=1,\ldots,[\log_2(n)], j<2^{i+1}$, where $M'_{i}(\X_j)$ is the approximation for the sample $j$ based on the first $2^i$ samples. Then, the number of predictions $M'_{i}(\X_j)$ is not larger than $\sum_{0 \le i \le \log_2(n)} 2^{i+1} < 4n$. The gradient on the example $X_k$ used for choosing a tree structure is estimated on the basis of the approximation $M'_i(X_k)$, where $i=[\log_2(k)]$.

\section{Fast scorer}\label{sec:calc}
CatBoost uses oblivious trees as base predictors. In such trees the same splitting criterion is used across an entire level of the tree~\cite{kohavi1995oblivious,langley1994oblivious}. Such trees are balanced and less prone to overfitting.
Gradient boosted oblivious trees were successfully used in various learning tasks~\cite{ferov2016enhancing,gulin2011winning}.
In oblivious trees each leaf index can be encoded as a binary vector with length equal to the depth of the tree. This fact is widely used in CatBoost model evaluator: we first binarize all used float features, statistics and one-hot encoded features and then use binary features to calculate model predictions.

All binary feature values for all examples are stored in a continuous vector $B$. Leaf values are stored in a float vectors of size $2^{d}$, where $d$ is the tree depth.
To calculate the leaf index for the $t$-th tree and for an example $x$ we build a binary vector $\sum_{i = 0}^{d-1}{2^{i} \cdot B(x, f(t, i))}$, where $B(x,f)$ is the value of the binary feature $f$ on the example $x$ that we read from the vector $B$ and $f(t, i)$ is the number of the binary feature from $t$-th tree on depth $i$.

That vectors can be built in a data parallel manner which gives up to 3x speedup.
This results in a much faster scorer than all existing ones as shown in our experiments.

\section{Fast training on GPU}\label{Sec:gpu}

\paragraph{Dense numerical features}
One of the most important building blocks for any GBDT implementation is searching for the best split. This block is the main computation burden for building decision tree on dense numerical datasets. CatBoost uses oblivious decision trees as base learners and performs feature discretization into a fixed amount of bins to reduce memory usage~\cite{gulin2011winning}. The number of bins is the parameter of the algorithm.
As a result we could use histogram-based approach for searching for best splits.
Our approach to building decision trees on GPU is similar in spirit to one described in~\cite{lightgbmGPU}. 
We group several numerical features in one 32-bit integer and currently use:
\begin{itemize}[noitemsep,nolistsep]
\item 1 bit for binary features and group 32 features per integer.
\item 4 bits for features with no more than 15 bins, 8 features per integer.
\item 8 bits for other features (maximum feature discretization is 255), 4 features per integer.
\end{itemize}

In terms of GPU memory usage CatBoost is at least as efficient as LightGBM~\cite{lightgbmGPU}. The main difference is another way of histogram computation. Algorithms in LightGBM and XGBoost\footnote{\url{https://devblogs.nvidia.com/parallelforall/gradient-boosting-decision-trees-xgboost-cuda/}} have a major drawback: they rely on atomic operations. Such technique is very easy to deal with concurrent memory accesses but it is also relatively slow even on the modern generation of GPU.
Actually histograms could be computed more efficiently without any atomic operations. We will describe here only the basic idea of our approach by a simplified example: simultaneous computation of four 32-bin histograms with a single float additive statistic per feature. This idea could be efficiently generalized for cases with several statistics and multiple histograms.

So we have gradient values $g[i]$ and feature groups $(f_1, f_2, f_3, f_4)[i]$.
We need to compute 4 histograms:
$
\text{hist}[j][b] = \sum_{i: f_j[i] = b} g[i] 
$.
CatBoost builds partial histogram per each warp\footnote{group of $32$ threads working in parallel on current NVIDIA GPU} histograms instead of histogram per thread block. We will describe work which is done by one warp on first 32 samples. Thread with index $i$ processes sample $i$. Since we are building $4$ histograms at once we need $32 * 32 * 4$ bytes of shared memory per warp. To update histograms all $32$ threads load sample labels and grouped features to registers. Then warp performs updates of shared-memory histogram simultaneously in $4$ iterations: on $l$-th ($l=0\ldots 3$)
iteration thread with index $i$  works with feature $f_{(l + i) \text{ mod } 4}$ and adds $g[i]$ for $\text{hist}[(l + i) \text{ mod } 4][f_{(l + i) \text{ mod } 4}]$. With proper histogram layout this operation could avoid any bank conflicts and add statistics via all 32 threads in parallel. 

CatBoost implementation builds histograms for 8 features per group and 2 statistics; for 32 binary features per group and 2 statistics; for 4 features per group and 2 statistics with bin count 32, 64, 128 and 255. In order to achieve fast computation of all these histograms we have to use all available shared memory. As a result our code can't achieve 100\% occupancy.
So we do loop unrolling to utilize instruction-level parallelism. This technique allows high performance even at lower occupancy\footnote{\url{http://www.nvidia.com/content/GTC-2010/pdfs/2238_GTC2010.pdf}}

\paragraph{Categorical features}
\label{subsec:cat_features_gpu}

CatBoost implements several ways to deal with categorical features. For one-hot encoded features we don't need any special treatment~--- histogram-based approach for split searching can be easily adopted to such case. Statistics computation for single categorical features also could be done during preprocessing stage. CatBoost also uses statistics for feature combinations. Dealing with them is the slowest and most memory consuming part of algorithm.

We use perfect hash to store values of categorical features to reduce memory usage. Because of memory constraints on GPU, we store bit-compressed perfect hashes in CPU RAM and stream required data on demand, overlapping computation and memory operations.
Construction of feature combinations on the fly requires us to dynamically build (perfect) hash functions for this new features and compute statistics with respect to some permutation (via \ref{eq:ordered_ctr}) for each unique value of hash. We use radix sort to build perfect hashes and group observations by hash. In every group we need to compute prefix sums of some statistic. Computation of this statistics is done with segmented scan GPU primitive (CatBoost segmented scan implementation is done via operator transformation~\cite{segScan} and is based on highly-efficient implementation of scan primitive in CUB~\cite{scanCub}).
 
\paragraph{Multiple GPU support}

CatBoost GPU implementation supports several GPUs out of the box. Distributed tree learning could be parallelized by samples or by features.
CatBoost uses a computation scheme with several permutations of learning dataset and computes statistics for categorical features during training. So we need to use feature parallel learning.   

\section{Experiments}\label{Sec:Experiments}
\paragraph{Quality: comparison with baselines}

We compare our algorithm with XGBoost, LightGBM and H2O. The results of the comparison are presented in Table~\ref{tab:baselines1}. The detailed description of the experimental setup as well as dataset descriptions are published on our github together with the code of the experiment and a container with all the used libraries, so that the result can be reproduced. In this comparison we made categorical features preprocessing using the statistics on random permutation of data. The parameter tunning and training was performed on 4/5 of the data and the testing was performed on the other 1/5. The training with selected parameters was performed 5 times with different random seeds, the result is the average logloss on 5 runs.

The table shows that the CatBoost algorithm outperforms other algorithms on all datasets in classification task. In our repo on github you can also see that CatBoost with default parameters outperforms tunned XGBoost and H2O on all datasets and LightGBM on all but one datasets.

\begin{table}[t]
\tiny
  \caption{Comparison with baselines. Tuned algorithms. Logloss}
  \label{tab:baselines1}
  \vspace{-1pt}
  \centering
  \begin{tabular}{lllll}
    \cmidrule{2-5}
   & CatBoost &  LightGBM &  XGBoost & H2O \\
     \midrule
Adult & \textbf{0.269741}  & 0.276018  (+2.33\%) & 0.275423 (+2.11\%) & 0.275104 (+1.99\%) \\
Amazon & \textbf{0.137720} & 0.163600 (+18.79\%) & 0.163271 (+18.55\%) & 0.162641 (+18.09\%) \\
Appet & \textbf{0.071511} & 0.071795 (+0.40\%) & 0.071760 (+0.35\%) & 0.072457 (+1.32\%) \\
Click & \textbf{0.390902} & 0.396328 (+1.39\%) & 0.396242 (+1.37\%) & 0.397595 (+1.71\%) \\
Internet & \textbf{0.208748} & 0.223154  (+6.90\%) & 0.225323  (+7.94\%) & 0.222091  (+6.39\%) \\
Kdd98 & \textbf{0.194668}  & 0.195759 (+0.56\%) & 0.195677 (+0.52\%) & 0.195395 (+0.37\%) \\
Kddchurn & \textbf{0.231289}  & 0.232049  (+0.33\%) & 0.233123  (+0.79\%) & 0.232752  (+0.63\%) \\
Kick & \textbf{0.284793}  & 0.295660 (+3.82\%) & 0.294647 (+3.46\%) & 0.294814  (+3.52\%)
\\
\bottomrule
\end{tabular}
    \vspace{-17pt}
\end{table}

\paragraph{GPU vs CPU training performance}
\begin{wraptable}{r}{5.5cm}
\vspace{-9pt}
\tiny
  \caption{GPU vs CPU training}
  \label{tab:gpu_cpu}
  \centering
  \vspace{-9pt}
  \begin{tabular}{llll}
    \toprule
    &  \multicolumn{2}{c}{Epsilon} & \multicolumn{1}{c}{Criteo}      \\
    \cmidrule(r){2-3}
    \cmidrule{4-4}
     & 128 bins & 32 bins  &  128 bins  \\
    \midrule
    CPU &  713  $(1.0)$ & 653 $(1.0)$ & 1060 ($1.0$) \\
    K40  & 547 ($1.3$) & 248 $(2.6)$  &  373 ($2.84$)\\
    GTX 1080&    194 $(3.67)$ & 120 $(5.4)$  &      285 ($3.7$) \\
    P40  &  162 $(4.4)$ & 91 $(7.1)$  & 123 ($8.6$)\\
    GTX 1080Ti  &  145 $(4.9)$ & 88 $(7.4)$ &  301 ($3.5$) \\
    P100-PCI  &  127 $(5.6)$ & 70 $(9.3)$ &   82 ($12.9$) \\
    V100-PCI  &  77 $(9.25)$ &  49 $(13.3)$ &  69.8 ($15$) \\
    \bottomrule
  \end{tabular}
  \vspace{-15pt}
\end{wraptable}
Scripts for running GPU experiments are in our github repo. In the first experiment we compared training speed for our GPU vs CPU implementations. For CPU version we used dual-socket server with 2 Intel~Xeon CPU (E5-2650v2,~2.60GHz) and 256GB RAM and run CatBoost in 32 threads (equal to number of logical cores).  GPU implementation was run on several servers with different GPU types. Our GPU implementation doesn't require multi-core server for high performance, so different CPU and 
machines shouldn't significantly affect 
benchmark results.

Results are present in table~\ref{tab:gpu_cpu}.
We used Criteo dataset ($36 * 10^6$ samples, $26$ categorical, $13$ numerical features) to benchmark our categorical features support. We used 2 GTX1080 because 1 had not enough memory. It's clearly seen that the GPU version significantly outperforms CPU training time even on old generation GPU (Tesla~K40) and gains impressive x15 speedup on 
NVIDIA~V100 card.

We used Epsilon dataset ($4 * 10^5$ samples, 2000 features) to benchmark our performance on dense  numerical dataset. For dense numerical dataset CatBoost GPU training time depends on level of feature discretization. In the table we report time for default 128 bins and for 32 bins which is often sufficient. We would like to mention that Epsilon dataset has not enough samples to fully utilize GPU, and with bigger datasets we observe larger speedups.

\vspace{-5pt}
\paragraph{GPU training performance: comparison with baselines}
\label{par:gpu_vs_baselines}

Its very hard to compare different boosting libraries in terms of training speed. Every library has a vast number of parameters which affect training speed, quality and model size in a non-obvious way. Every library has its unique quality/training speed trade-off's and can't be compared without domain knowledge (e.g. is $0.5\%$ of quality metric worth it to train model 3-4 times slower?). Plus for each library it is possible to obtain almost the same quality with different ensemble sizes and parameters. As a result, we can't compare libraries by time we need to obtain certain level of quality.

So we could give only some insights of how fast our GPU implementation could train a model of fixed size. We use Epsilon dataset ($4 * 10^5$ samples for train, $10^5$ samples for test). For this dataset we measure mean tree construction time one can achieve without using feature subsampling and/or bagging by CatBoost and 2 open-source implementations of boosting with GPU support:  XGBoost (we use histogram-based version, exact version is very slow) and LightGBM. We run all experiments on the same machines with NVIDIA P100 accelerator, dual-core Intel Xeon E5-2660 CPU and 128GB RAM.  For XGBoost and CatBoost we use default tree depth equal to 6, for LightGBM we set leafs count to 64 to have more comparable results. We set bin to 15 for all 3 methods. Such bin count gives the best performance and  the lowest memory usage for LightGBM and CatBoost (Using 128-255 bin count usually leads both algorithms to run 2-4 times slower). For XGBoost we could use even smaller bin count but performance gains compared to 15 bins are too small to account for. All algorithms were run with 16 threads, which is equal to hardware core count. By default CatBoost uses bias-fighting scheme describe in section~\ref{sec:fighting}. This scheme is by design 2-3 times slower then classical boosting approach. GPU implementation of CatBoost contains a mode based on classic scheme for those who need best training performance, in this benchmark we used classic scheme.

\begin{figure}
\centering
  \caption{GPU learning curves}
  \label{fig:learning_curves}
  \vspace{-7pt}
\begin{subfigure}{.5\textwidth}
  \centering
  \includegraphics[width=.95\linewidth]{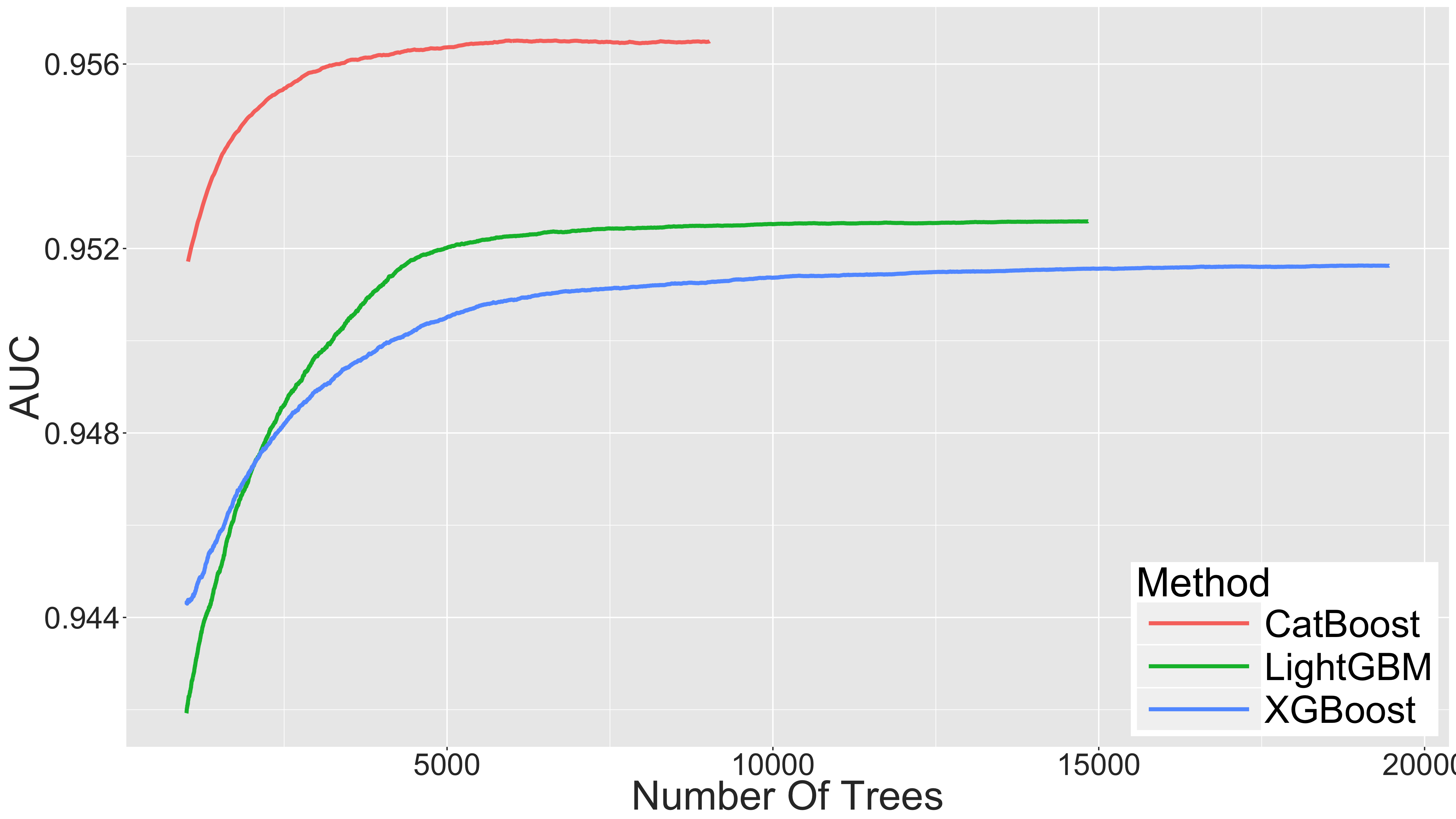}
  \caption{AUC vs Number of trees}
  \label{fig:auc_vs_iteration}
\end{subfigure}%
\begin{subfigure}{.5\textwidth}
  \centering
  \includegraphics[width=.95\linewidth]{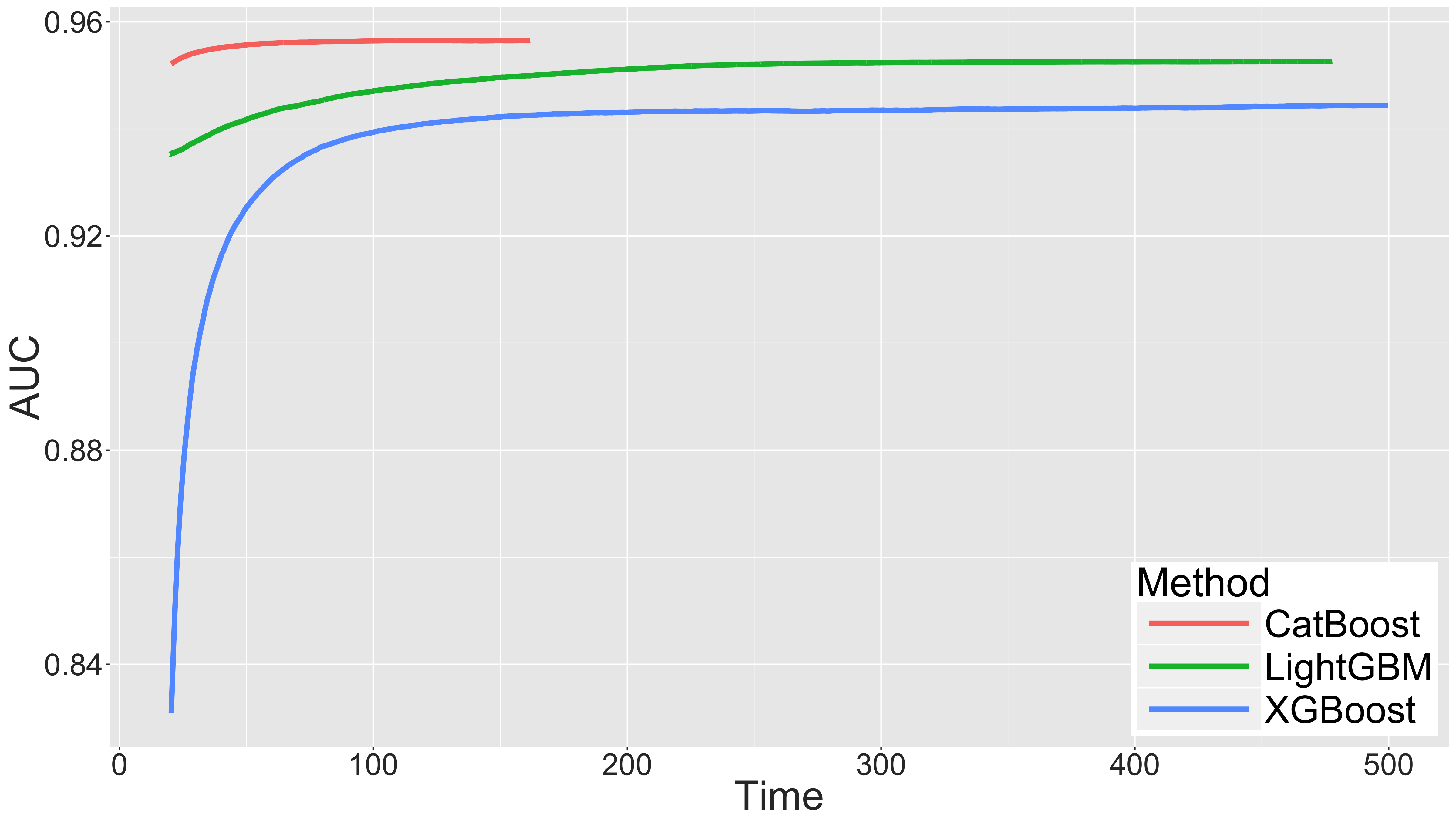}
  \caption{AUC vs Time}
  \label{fig:auc_vs_time}
\end{subfigure}
\vspace{-17pt}
\end{figure}

We set such learning rate that algorithms start to overfit approximately after 8000 trees (learning curves are displayed at figure~\ref{fig:learning_curves}, quality of obtained models differs by approximately $0.5\%$). We measure time to train ensembles of 8000 trees. Mean tree construction time for CatBoost was 17.9ms, for XGBoost 488ms, for LightGBM 40ms. These times are very rough speed comparison, because training time of one tree construction depends on distribution of features and ensemble size. At the same times it shows that if we have similar size ensembles we could expect CatBoost and LightGBM to be competitors for the fastest method, while XGBoost is significantly slower than both of them.

\vspace{-5pt}
\paragraph{Scorer performance}
\begin{wraptable}{r}{4.5cm}
\vspace{-10pt}
\tiny
  \caption{Scorer comparison}
  \label{tab:apply}
  \vspace{-1pt}
  \centering
  \begin{tabular}{lll}
    \cmidrule{2-3}
   & 1 thread &  32 threads\\
     \midrule
CatBoost & 2.4s  & 231ms \\
XGBoost & 78s (x32.5) & 4.5s (x19.5) \\
LightGBM & 122s (x50.8) & 17.1s (x74) \\
\bottomrule
\end{tabular}
    \vspace{-10pt}
\end{wraptable}

We used LightGBM, XGBoost and CatBoost models for Epsilon dataset trained as described above.
For each model we limit number of trees used for evaluation to $8000$ to make results comparable for the reasons described above. Thus this comparison gives only some insights of how fast the models can be scored.
For each algorithm we loaded test dataset in python, converted it to the algorithm internal representation and measured wall-time of model predictions on Intel Xeon E5-2660 CPU with 128GB RAM. The results are present in Table~\ref{tab:apply}. We can see that on similar sizes of ensembles CatBoost can be scored around 25 times faster than XGBoost and around 60 times faster than LightGBM.

\bibliography{boosting}

\begin{thebibliography}{10}

\bibitem{breiman1996out}
L.~Breiman.
\newblock Out-of-bag estimation, 1996.

\bibitem{caruana2006empirical}
R.~Caruana and A.~Niculescu-Mizil.
\newblock An empirical comparison of supervised learning algorithms.
\newblock In {\em Proceedings of the 23rd international conference on Machine
  learning}, pages 161--168. ACM, 2006.

\bibitem{cestnik1990estimating}
B.~Cestnik et~al.
\newblock Estimating probabilities: a crucial task in machine learning.
\newblock In {\em ECAI}, volume~90, pages 147--149, 1990.

\bibitem{chen2016xgboost}
T.~Chen and C.~Guestrin.
\newblock Xgboost: A scalable tree boosting system.
\newblock In {\em Proceedings of the 22Nd ACM SIGKDD International Conference
  on Knowledge Discovery and Data Mining}, pages 785--794. ACM, 2016.

\bibitem{dorogush2017fighting}
A.~V. Dorogush, A.~Gulin, G.~Gusev, L.~Ostroumova~Prokhorenkova, and
  A.~Vorobev.
\newblock Catboost: unbiased boosting with categorical features.
\newblock {\em arXiv preprint arXiv:1706.09516}, 2017.

\bibitem{scanCub}
M.~G. Duane Merrill NVIDIA~Corporation.
\newblock Single-pass parallel prefix scan with decoupled look-back.
\newblock Technical report, NVIDIA, 2016.

\bibitem{ferov2016enhancing}
M.~Ferov and M.~Modr{\`y}.
\newblock Enhancing lambdamart using oblivious trees.
\newblock {\em arXiv preprint arXiv:1609.05610}, 2016.

\bibitem{friedman2001greedy}
J.~H. Friedman.
\newblock Greedy function approximation: a gradient boosting machine.
\newblock {\em Annals of statistics}, pages 1189--1232, 2001.

\bibitem{friedman2002stochastic}
J.~H. Friedman.
\newblock Stochastic gradient boosting.
\newblock {\em Computational Statistics \& Data Analysis}, 38(4):367--378,
  2002.

\bibitem{gulin2011winning}
A.~Gulin, I.~Kuralenok, and D.~Pavlov.
\newblock Winning the transfer learning track of yahoo!'s learning to rank
  challenge with yetirank.
\newblock In {\em Yahoo! Learning to Rank Challenge}, pages 63--76, 2011.

\bibitem{lightgbmGPU}
C.-J.~H. Huan~Zhang, Si~Si.
\newblock Gpu-acceleration for large-scale tree boosting.
\newblock {\em arXiv:1706.08359}, 2017.

\bibitem{kohavi1995oblivious}
R.~Kohavi and C.-H. Li.
\newblock Oblivious decision trees, graphs, and top-down pruning.
\newblock In {\em IJCAI}, pages 1071--1079. Citeseer, 1995.

\bibitem{langley1994oblivious}
P.~Langley and S.~Sage.
\newblock Oblivious decision trees and abstract cases.
\newblock In {\em Working notes of the AAAI-94 workshop on case-based
  reasoning}, pages 113--117. Seattle, WA, 1994.

\bibitem{micci2001preprocessing}
D.~Micci-Barreca.
\newblock A preprocessing scheme for high-cardinality categorical attributes in
  classification and prediction problems.
\newblock {\em ACM SIGKDD Explorations Newsletter}, 3(1):27--32, 2001.

\bibitem{roe2005boosted}
B.~P. Roe, H.-J. Yang, J.~Zhu, Y.~Liu, I.~Stancu, and G.~McGregor.
\newblock Boosted decision trees as an alternative to artificial neural
  networks for particle identification.
\newblock {\em Nuclear Instruments and Methods in Physics Research Section A:
  Accelerators, Spectrometers, Detectors and Associated Equipment},
  543(2):577--584, 2005.

\bibitem{segScan}
M.~G. Shubhabrata~Sengupta, Mark~Harris.
\newblock Efficient parallel scan algorithms for gpus.
\newblock Technical report, NVIDIA, 2008.

\bibitem{wu2010adapting}
Q.~Wu, C.~J. Burges, K.~M. Svore, and J.~Gao.
\newblock Adapting boosting for information retrieval measures.
\newblock {\em Information Retrieval}, 13(3):254--270, 2010.

\bibitem{zhang2015gradient}
Y.~Zhang and A.~Haghani.
\newblock A gradient boosting method to improve travel time prediction.
\newblock {\em Transportation Research Part C: Emerging Technologies},
  58:308--324, 2015.

\end{thebibliography}

\bibliographystyle{abbrv}

\end{document}